\documentclass[10pt,conference,letterpaper]{IEEEtran}
\usepackage{times,amsmath,epsfig}
\usepackage{dsfont}
\usepackage{amssymb}
\usepackage{graphicx}
\usepackage{dirtytalk}
\usepackage{mathtools}
\usepackage{rotating}
\usepackage{multirow}

\usepackage{tikz}

\usepackage[caption=false]{subfig}
\usepackage[export]{adjustbox}
\usepackage{array}

\usepackage[pagebackref=true,breaklinks=true,colorlinks,bookmarks=false]{hyperref}

\newcommand{\iqry}{{q}}

\newcommand{\ipos}{{p}}

\newcommand{\ineg}{{n}}

\newcommand{\loss}{\ell}

\title{Images \& Recipes: Retrieval in the cooking context}

\IEEEoverridecommandlockouts
\begin{document}
    \author{%
        {Micael Carvalho{\small $~^{1 *}$}\thanks{$^*$ Equal contribution}, Remi Cadene{\small $~^{1 *}$}, David Picard{\small $~^{1,2}$}, Laure Soulier{\small $~^{1}$}, Matthieu Cord{\small $~^{1}$}}
        \vspace{1.6mm}\\
        \fontsize{10}{10}\selectfont\itshape
        $^{1}$\,Sorbonne Universit\'e -- Paris, France
        \vspace{1.6mm}\\
        \fontsize{10}{10}\selectfont\rmfamily\itshape
        $^{2}$\,ETIS, UMR 8051 -- Universit\'e Paris Seine, Universit\'e Cergy-Pontoise, ENSEA, CNRS\\
        \\
        An extended/complete version of this work is to appear at SIGIR 2018, and is available at \url{https://arxiv.org/abs/1804.11146}
    }
    
    \maketitle
    
    \begin{abstract}
    
            Recent advances in the machine learning community allowed different use cases to emerge, as its association to domains like cooking which created the computational cuisine. 
            In this paper, we tackle the picture-recipe alignment problem, having as target application the large-scale retrieval task (finding a recipe given a picture, and vice versa).
            Our approach is validated on the Recipe1M dataset, composed of one million image-recipe pairs and additional class information, for which we achieve state-of-the-art results.
           
    \end{abstract}
    
    \section{Introduction}
    Cooking is one of the most fundamental human activities connected to various aspects of human life such as food, health, dietary, culinary art, and so on.
Data mining and machine learning techniques have been used to extract and clean large datasets of recipes from the Internet, and also to plan and analyze the recipe instructions. One difficulty underlying computational cooking relies on the nature of data since recipes generally include images and text, whether structured or unstructured (e.g., the list of ingredients or instructions  in natural language).  This opens several challenges in terms of indexing/storing and gives rise to numerous application tasks, such as recommendation or classification.  Computational cooking has consequently emerged as a new research topic that also benefits from recent advances in machine learning based on deep neural approaches. More particularly, these approaches aim at projecting data elements in a latent semantic space so as similar elements are represented with similar low-dimensional representations~\cite{Harris54,mikolov2013distributed}. Beyond solving indexing issues, these latent representations, also called \textit{"embeddings"}, allow machines to perceive texts and images in a meaningful way, similar to that of humans, that could be exploited in smart cooking-oriented tasks, such as ingredient identification~\cite{chen2016}, recipe recognition~\cite{wang2015}, or recipe popularity prediction~\cite{SanjoK17}.

In this paper, we are interested in smart retrieval between recipe component modalities (namely recipe texts and cooked dish pictures) in the cooking context. We propose a deep neural framework aiming at jointly learning  the representation of recipe component modalities (namely recipe texts and cooked dish pictures) and retrieving relevant pictures of a meal given its recipe or, conversely, a relevant recipe given an image query.
More particularly, we carry out two deep neural networks to embed both recipes and images into a common semantic representation space. The neural networks are trained such that the representation of a dish picture and the representation of its corresponding recipe are similar, whereas non-corresponding images and recipes have highly dissimilar representations. This refers to as \textit{"cross-modal alignment"}.
To train these neural networks, we use the very recently released dataset Recipe1M~\cite{salvador2017}, composed of one million pairs of aligned text documents and images corresponding to cooking recipes with matching pictures.

The contribution of our paper is twofold:
\begin{itemize}
    \item First, we propose a deep neural model that  learns the representation of recipe texts and images relying on a multi-modal retrieval learning objective function (Section \ref{sec:model}). This model is evaluated in Section \ref{sec:exps}.
    \item Second, we discuss the potential of such model for solving traditional and promising computational cooking use-cases. We analyze several downstream tasks to exploit the proposed model (in terms of learned architecture and/or representations) and qualitatively demonstrate its effectiveness (Section \ref{sec:usecase}).
\end{itemize}

\begin{center}
    \begin{table*}[t]
        \centering
        \begin{tabular}{m{.1cm}m{4cm}m{8cm}c}
            \multicolumn{4}{c}{\includegraphics[width=\linewidth]{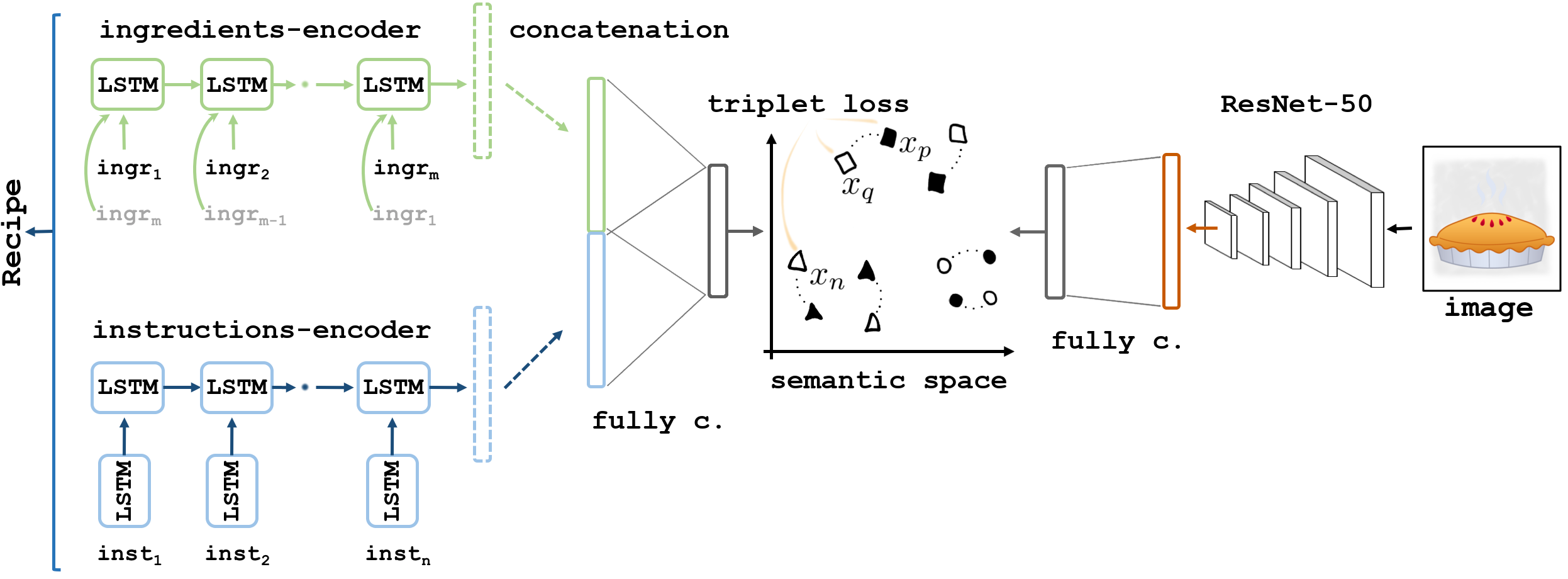}}\\
            & & & \\
            & \multicolumn{1}{c}{\textbf{ingr (ingredients)}} & \multicolumn{1}{c}{\textbf{instr (cooking instructions)}} & \multicolumn{1}{c}{\textbf{image}} \\
            \normalsize\begin{turn}{90}Pizza\end{turn} &
        	{\small
        	    \textit{
        	        \begin{enumerate}
                	    \item pizza dough
                	    \item hummus
                	    \item arugula
                	    \item cherry / grape tomatoes
                	    \item pitted greek olives
                	    \item crumbled feta cheese
        	        \end{enumerate}
        	   }
        	} &
            {\small
                \textit{
        	        \begin{enumerate}
                        \item Cut the dough into two 8-ounce sized pieces.
                        \item Roll the ends under to create round balls.
                        \item Then using a well-floured rolling pin, roll the dough out into 12-inch circles.
                        \item Place the dough circles on sheets of parchment paper.
                        \item[...] ...
        	        \end{enumerate}
                }
            } &
            \raisebox{-.5\height}{\includegraphics[width=.2\linewidth]{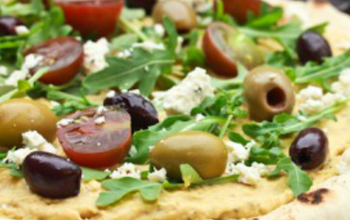}}\\
            \normalsize\begin{turn}{90}Pecan Pie\end{turn} &
            {\small
        	    \textit{
        	        \begin{enumerate}
                	    \item unsalted butter
                	    \item eggs
                	    \item condensed milk
                	    \item sugar
                	    \item vanilla extract
                	    \item chopped pecans
                	    \item chocolate chips
                	    \item[...] ...
        	        \end{enumerate}
        	   }
        	} &
            {\small
        	    \textit{
        	        \begin{enumerate}
                	    \item Preheat the oven to 375 degrees F.
                	    \item In a large bowl, whisk together the melted butter and eggs until combined.
                	    \item Whisk in the sweetened condensed milk, sugar, vanilla, pecans, chocolate chips, butterscotch chips, and coconut.
                	    \item[...] ...
        	        \end{enumerate}
        	   }
        	} &
            \raisebox{-.5\height}{\includegraphics[width=0.2\linewidth]{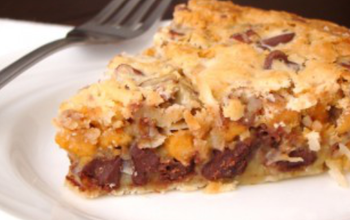}}\\
        \end{tabular}
        \vspace{\baselineskip}
        \vspace{-0.4cm}
        \caption{\small Overview of our multi-modal retrieval neural network (on the top) and examples of inputs issued from the large-scale Recipe1M dataset (on the bottom).  
        }
        \label{fig:our_architecture}
        \vspace{-0.8cm}
    \end{table*}
\end{center}

    \section{Related work}
    Smart cooking has recently become the center of increased interest as shown by the development of related workshops such as the Workshop on Multimedia for Cooking and Eating Activities~\cite{cea2017}.
In this domain, the increasing importance of food-related tasks in computer vision has motivated the creation of many related datasets. As a first example, \cite{chenpfid2009icip} proposed the Pittsburgh fast-food image dataset, containing 4,556 pictures of fast-food plates. Other examples are the datasets proposed by \cite{Kawano2014}, which contains approximately 10,000 images of 100 mainly-Japanese food categories, and by \cite{Farinella2015}, containing 889 distinct plates.

In order to solve more complex tasks, other initiatives provided richer sets of images. For example, \cite{bossard2014} proposed the Food-101 dataset, containing around 101,000 images of 101 different categories. In the same spirit, \cite{wang2015} introduced a twin dataset, containing the same categories, but with recipes associated to each picture. Finally, the dataset proposed by \cite{chen2016} is similar in the number of images (110,241), while also including relations to 353 ingredients and 65,284 recipes.
Other ideas, like the one of \cite{Beijbom2015}, involve taking advantage of the GPS data, used to retrieve nearby restaurants and to match a picture, taken by the user, to items from the menu. Important information, like nutritional values, are then recovered.

More recently, the first very large scale dataset for food-related tasks was presented by \cite{salvador2017}. They collected nearly 1 million recipes, with about 800,000 images associated to them.  The goal of this dataset is to tackle the problem of cross-modal retrieval in which the user wants to retrieve images of a dish by providing its recipe or, conversely, retrieving a recipe by providing a picture of the meal. The authors introduce a dual neural network to align textual and visual representations in a common vector space.
They also provide high level semantic information, such as food classes that allow us to enforce a semantic structure on the learned representations (\textit{i.e.}, items from the same class to have similar representations).

To take full advantage of the large-scale aspect of the dataset proposed in \cite{salvador2017}, modern machine learning methods such as deep learning~\cite{Goodfellow-et-al-2016} can be used.
In particular, deep convolutional neural network~\cite{krizhevsky2012imagenet} are very popular to learn image representations, while word embedding~\cite{mikolov2013distributed} and recurrent neural networks~\cite{Hochreiter1997} are often used for text embedding.
To match visual and textual representation, metric learning is used to structure the representation space in such way that similar items (text or images) are close while dissimilar items are far away according to a learned distance measure in~that~space.

In this paper, we propose a deep architecture similar to~\cite{salvador2017} except that we do not mix the embedding framework with an additional classification branch. We introduce a completely different learning scheme. We argue that the pairwise approach of~\cite{salvador2017} is not fully appropriated for retrieval tasks. Instead, we propose a ranking-based scheme where similar items have a distance closer than that of dissimilar items. We then derive a general training scheme using triplet-based constraints, as integrated in 
the Large Margin Nearest Neighbor strategy \cite{weinbergerlmnn2009}. Triplet-based strategies have been successfully used to learn Visual Semantic Embeddings (VSE)~\cite{kiros2015unifying} with applications to captioning as a text retrieval task~\cite{karpathy2015deep}.
We first show that our learning scheme leads to outstanding results for retrieval tasks. We then illustrate some powerful applications of such a deep architecture for cooking purposes.

    \section{Proposed architecture}
    \label{sec:model}
    Our global architecture is depicted in \autoref{fig:our_architecture}. 
It consists of two branches based on deep neural networks that respectively map each modality (image or text recipe) into a common representation semantic space  in which they can be compared.

The image branch (top-right part of \autoref{fig:our_architecture}) is composed of a ResNet-50 model \cite{He2015}. 
It contains a total of 50 convolutional layers, totaling more than 25 million parameters.
This architecture is further detailed in~\cite{He2015}, and was chosen in order to obtain comparable results to \cite{salvador2017} by sharing a similar setup.
The ResNet-50 is pretrained on the large-scale dataset of the ImageNet Large Scale Visual Recognition Challenge~\cite{ILSVRC15}, containing 1.2 million images, and is fine-tuned with the whole architecture. 
This model is followed by a fully connected layer (indicated by \textit{fully c.} on \autoref{fig:our_architecture}), which maps the outputs of the network into the semantic space, and is trained from scratch.

In the recipe branch (top-left part of \autoref{fig:our_architecture}), ingredients and instructions are first embedded separately, and their obtained representations are then concatenated as input of a fully connected layer that maps the recipe features into the semantic space. For ingredients, we use a bidirectional LSTM~\cite{Hochreiter1997} on their  pretrained embeddings obtained with the word2vec algorithm~\cite{mikolov2013distributed}. With the objective to consider the different granularity levels of  the instruction text, we use a hierarchical LSTM in which the word-level is pretrained using the skip-thoughts technique~\cite{kiros2015} and is not fine-tuned while the sentence-level is learned from scratch.

The semantic space can be structured by optimizing the network parameters $\theta$ of both branches following a specific objective. It includes all of the weights of the transformations applied by the layers of our model.
In our case, we are interested in retrieval tasks and as such, we want a semantic space such that given a query $x_q$ (either recipe or image), any positive item $x_p$ (recipe or image) with respect to that query to be closer to $x_q$ than any negative item $x_n$ (recipe or image) with respect to the query. This distance is measured in the semantic space by $d_\theta(x_q, x_p)$ (respectively $d_\theta(x_q, x_n)$) and depends on the parameters $\theta$.
Ideally, we want to enforce a margin $\alpha$ between the distance with the positive items and the distance with the negative items, making sure that positive items are retrieved first (see \cite{weinbergerlmnn2009}), or more formally $d_\theta(x_q, x_p) + \alpha \leq d_\theta(x_q, x_n)$.
This is achieved by defining the following loss function that counts a penalty for each triplet $(x_q, x_p, x_n)$ that does not satisfy the previous inequality:
\begin{align}
    \label{eq:loss_tri}
    \loss_{tri}(\theta, x_\iqry, x_\ipos, x_\ineg) &= \text{max}(0,d_\theta(x_\iqry, x_\ipos) + \alpha - d_\theta(x_\iqry, x_\ineg))
\end{align}
Training the full architecture then consists in optimizing the parameters $\theta$ so as to minimize this loss function over all possible triplets $(x_q, x_p, x_n)$.

The structure enforced on the semantic space largely depends on the choice of triplets $(x_q, x_p, x_n)$.
In our case, we propose 2 different sources of triplets to leverage the multi-scale structure of semantic space (namely, fine-grained elements and semantic classes).
The first source is instance-based and considers matching pairs of image and the associated recipe for the positive items. This provides a fine-grained structure to the semantic space where the nearest item from the other modality (\emph{e.g.}, image given a recipe) with respect to the query is optimized to be the very same meal.
The second source is semantic and considers items that belong to the same semantic classes as given by the 1000 class labels provided in Recipe1M.
This enforces a semantic structure on the semantic space by making sure that related dishes are closer to each other than to non-related ones.
In practice, we only sample positively-related pairs (both instance based and semantic based), picking negatively-related counterparts inside the mini-batch after their evaluation.
This strategy allows us to reduce the computational cost of training the network, as we will only need two forward passes per triplet, instead~of~three.

    \section{Experiments}
    \label{sec:exps}
    \begin{table*}[t]
    \centering
            \begin{tabular}{ccccccccc}
                \hline
                & \multicolumn{4}{c}{image to recipe} & \multicolumn{4}{c}{recipe to image} \\ \cline{2-9}
                & MedR & R@1 & R@5  & R@10 & MedR & R@1 & R@5 & R@10 \\
                \hline
                CCA \cite{salvador2017} & 15.7 & 14.0 & 32.0 & 43.0 & 24.8 & 9.0 & 24.0 & 35.0 \\
                Im2Recipe \cite{salvador2017} & 5.2 & 24.0 & 51.0 & 65.0 & 5.1 & 25.0 & 52.0 & 65.0 \\
                Ours & $\mathbf{1.0} \pm 0.1$ & $\mathbf{39.8} \pm 1.8$ & $\mathbf{69.0} \pm 1.8$ & $\mathbf{77.4} \pm 1.1$ & $\mathbf{1.0} \pm 0.1$ & $\mathbf{40.2} \pm 1.6$ & $\mathbf{68.1} \pm 1.2$ & $\mathbf{78.7} \pm 1.3$ \\
                \hline
            \end{tabular}
    \vspace{\baselineskip}
    \caption{Comparison with the state of the art on image to recipe and recipe to image tasks. MedR stands for Median Rank. R@K means Recall at K in percent. Results over 10 bags of 1k pairs each.}
    \label{tab:sota}
    \vspace{-0.7cm}
\end{table*}

We conduct a quantitative evaluation of our proposal on the Recipe1M dataset \cite{salvador2017} including about 800,000 recipe-picture pairs. To learn the network parameter $\theta$, we use the standard splits of the dataset for training, validating, and testing~our~model.

Taking into account the retrieval objective of our model, we apply the following evaluation methodology: given a query (whether image or recipe), we rank a collection of 1000 items from the other modality by their increasing distance to the query in the semantic space. The objective is to rank in the best position the element associated to the query. 
For comparison purposes, we apply the same methodology on item representations learned from two state-of-the-art baselines: the classification-oriented neural network \cite{salvador2017} and the Canonical Correlation Analysis (CCA) \cite{hotelling1936relations}.

Two different metrics are used in our evaluation: (1) MedR, which stands for Median Rank, and evaluates the median position of the associated element in the ranking. A MedR of 0 indicates that the correct sample is always the best match in the semantic space. A random ranking would then have a theoretical MedR of 500, since each match is unique inside our tests. (2) The R@N, standing for Recall at N, indicates the percentage of queries where the relevant target instance is retrieved among the first N ranked items. For example, R@1 indicates the average number of correct matches found as the best match, while R@10 indicates the average number of searches where the correct match is found among the 10 closest  points in the semantic space.

\autoref{tab:sota} presents the obtained results. 
We observe that our scheme drastically improves over the current state-of-the-art models for this dataset. On both tasks of \textit{image to recipe} and \textit{recipe to image} we achieve a MedR (lower is better) of 1.0, while the best known results are of 5.2 and 5.1, respectively. A similar behavior can be observed when analyzing the recall metric (higher is better): we obtain 39.8 and 40.2 for R@1, while the best previous results were 24.0 and 25.0. The same conclusions can be extended to R@5 and R@10, for which our approach also overtakes the state-of-the-art.

In order to grasp the structure of the semantic space, we present examples of use cases in the following section.

    \section{Study of the semantic space}
    \label{sec:usecase}

    Our model has a twofold objective (namely, cross-modal retrieval and multi-modal representation) which could be beneficial for several cooking-related tasks.
    In this section, we discuss the potential of our model for promising cooking-related application tasks.
    The strength of our model relies on the fact that, on one hand, the network architecture allows to answer retrieval tasks, and on the other hand, the learned representation space allows to identify similar/dissimilar text and visual items. We believe that such systems might be beneficial for ambitious tasks, such as menu or shopping list generation, or calorie tracking. These tasks require more insight in terms of model design since they might include diversity factors, ingredient quantity analysis, or external knowledge in a task-oriented model. In what follows, we particularly focus on downstream tasks in which the current setting might be applied. We provide illustrative examples issued from the testing set of our evaluation process.
    For better readability, we always show the results as images, even for text recipes for which we display their corresponding~original~picture.

    \subsection{Multi-modal retrieval}
    The first task relies on multi-modal retrieval, for which a user requests items in any available format given a query item in a specific format. This results in image-to-text, text-to-image, text-to-text, and image-to-image retrieval scenarios, referred to as \textit{"multi-modal retrieval"}.
    In long term, this could be useful when the user needs the recipe of a meal eaten in a restaurant or identifying similar recipes if they would like to replace a meal in their menu.
    In our case, solving this task leads to retrieve the most similar items in the semantic space (\textit{i.e.}, items with the smallest distances).
    For illustrating our intuition, we test the four retrieval scenarios on the query shown in \autoref{tab:visu_modality-to-modality_query}.
    
    Regarding the \textit{image-to-image} scenario (see \autoref{tab:visu_modality-to-modality_image-to-image}),  
    we can see that the top retrieved images look similar to the query image not only in term of colors, shapes, and textures, but also semantically. For instance, the first, third and forth images have gratin cheese on top and the second image also has a plate which looks similar to the one from the query. 
    When looking at their corresponding recipe, all five include a similar set of ingredients containing potatoes, milk, butter, cheese, and onion. Small variations in the ingredients are observed, for example, the second image has rice instead of potatoes. As for the instructions, all shown results are baked in a 350 degrees Fahrenheit oven during 15 to 45 minutes depending~on~the~recipe.
    
    In the \textit{image-to-recipe} scenario (see \autoref{tab:visu_modality-to-modality_image-to-image}), 
    most of the results are shared with the image-to-image search which indicates that the embeddings of matching image-recipe pairs are very close.
    However, we also obtain results that are less visually similar, but are close to the recipe associated with the query, either in terms of ingredients or in cooking instructions.
    
    Starting from a text recipe query, \textit{recipe-to-image} in \autoref{tab:visu_modality-to-modality_image-to-image} shows the retrieved pictures. We are able to find images similar to the picture associated with the query recipe, although no visual information was used for the querying.
    Finally, the \textit{recipe-to-recipe} scenario (see \autoref{tab:visu_modality-to-modality_image-to-image}) highlights that although the ingredients and the cooking instructions of the retrieved recipes are similar to those of the query, we observe more visual diversity among the results.

    \begin{table*}[t]
    \centering
    \begin{tabular}{m{0.5cm}m{2.5cm}m{7.5cm}c}
        
        &
        \multicolumn{1}{c}{\textbf{Ingredients}}
        &
        \multicolumn{1}{c}{\textbf{Cooking instructions}} & \multicolumn{1}{c}{\textbf{Image}} \\
        
        \normalsize\begin{turn}{90}Crunchy Onion Potato Bake\end{turn}
        &
        {
            \small \textit{Milk, Water, Butter, Mashed potatoes, Corn, Cheddar cheese, French-fried onions}
        }
        &
        {
            \small \textit{
                Preheat oven to 350 degrees Fahrenheit.
                Spray pan with non stick cooking spray.
                Heat milk, water and butter to boiling; stir in contents of both pouches of potatoes; let stand one minute.
                Stir in corn.
                Spoon half the potato mixture in pan.
                Sprinkle half each of cheese and onions; top with remaining potatoes.
                Sprinkle with remaining cheese and onions.
                Bake 10 to 15 minutes until cheese is melted.
                Enjoy !
            }
        } &
        \raisebox{-.5\height}{\includegraphics[width=0.3\linewidth]{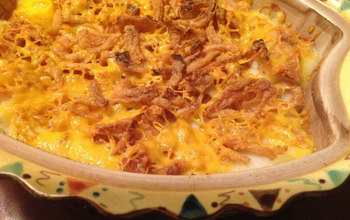}}

    \end{tabular}
    \vspace{\baselineskip}
    \vspace{-0.4cm}
    \caption{Query used in the multi-modal retrieval tasks.}
    \label{tab:visu_modality-to-modality_query}
    \end{table*}
    
    \begin{table*}[t]
        \centering
        \begin{tabular}{c c c c c c}
            
            & Top 2 & Top 3 & Top 4 & Top 5 & Top 6 \\
        
          \begin{turn}{90}image-to-image\end{turn} &
          \includegraphics[width=0.17\linewidth]{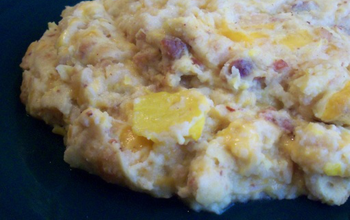} &
          \includegraphics[width=0.17\linewidth]{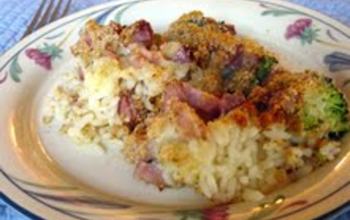} &
          \includegraphics[width=0.17\linewidth]{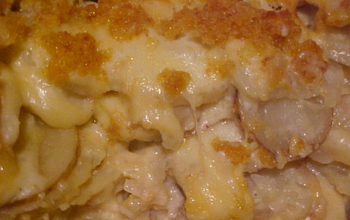} &
          \includegraphics[width=0.17\linewidth]{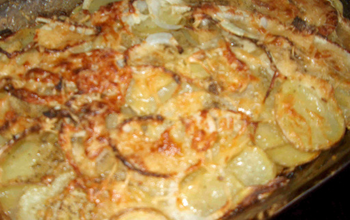} &
          \includegraphics[width=0.17\linewidth]{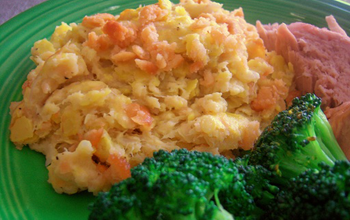} \\
          
          \begin{turn}{90}image-to-recipe\end{turn} &
          \includegraphics[width=0.17\linewidth]{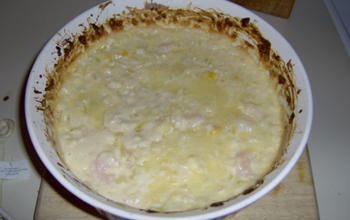} &
          \includegraphics[width=0.17\linewidth]{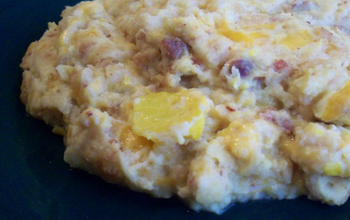} &
          \includegraphics[width=0.17\linewidth]{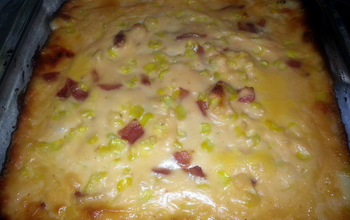} &
          \includegraphics[width=0.17\linewidth]{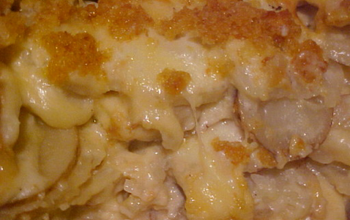} &
          \includegraphics[width=0.17\linewidth]{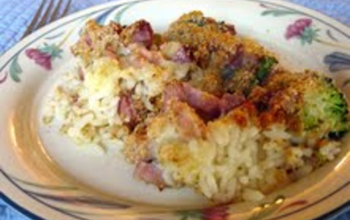} \\
          
          \begin{turn}{90}recipe-to-image\end{turn} &
          \includegraphics[width=0.17\linewidth]{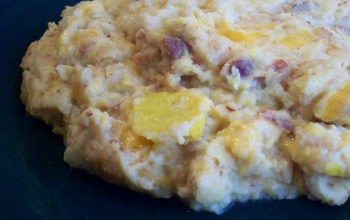} &
          \includegraphics[width=0.17\linewidth]{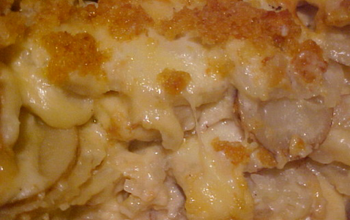} &
          \includegraphics[width=0.17\linewidth]{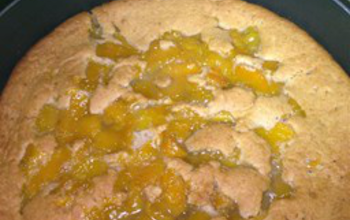} &
          \includegraphics[width=0.17\linewidth]{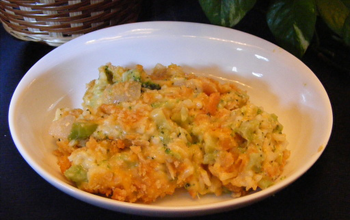} &
          \includegraphics[width=0.17\linewidth]{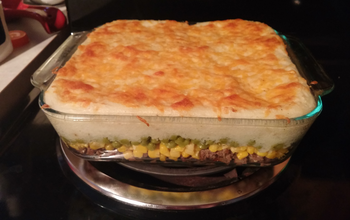} \\
          
          \begin{turn}{90}recipe-to-recipe\end{turn} &
          \includegraphics[width=0.17\linewidth]{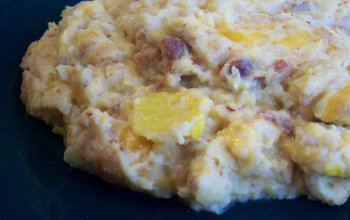} &
          \includegraphics[width=0.17\linewidth]{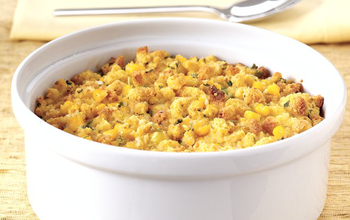} &
          \includegraphics[width=0.17\linewidth]{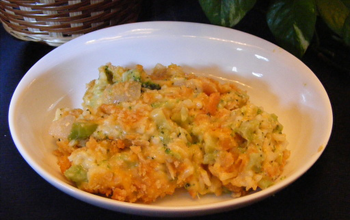} &
          \includegraphics[width=0.17\linewidth]{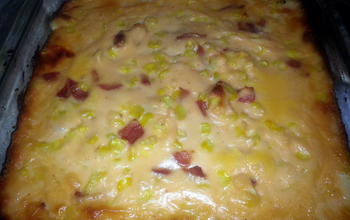} &
          \includegraphics[width=0.17\linewidth]{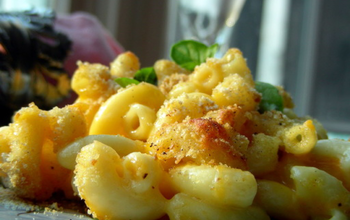}
        \end{tabular}
        \vspace{\baselineskip}
        \vspace{-0.4cm}
        \caption{Visualization of the top 2 to top 6 recipes (the image associated to recipe is displayed) for the multi-modal retrieval tasks. }
        \label{tab:visu_modality-to-modality_image-to-image}
        \vspace{-0.8cm}
    \end{table*}

    \subsection{Ingredient To Image}
    An interesting ability of our model is to map ingredients into the semantic space, in order to retrieve recipes containing the same ingredients. This is particularly useful when one would like to know what they can cook using aliments available in their fridge.
    
    To demonstrate this process, we create ingredient queries by averaging representation vectors of available ingredients, and then retrieve the nearest neighbors among 10,000 images randomly picked from the testing set. 
    In \autoref{tab:visu_ingr_to_image_mushroom}, we showcase images within the top 20 nearest neighbors of the ingredients \textit{carrot} and \textit{mushroom}.
    We are able to retrieve visually diverse meals containing the query ingredient.
    
    We show on \autoref{tab:visu_ingr_to_image_pizza} examples of retrieved images when searching for different ingredients while constraining the results to the class \emph{pizza}. Searching for \emph{pineapple} or \emph{olives} results in different types of pizzas. An interesting remark is that searching for \emph{strawberries} inside the class \emph{pizza} yields images of \emph{fruit pizza} containing strawberries, \textit{i.e.}, images that are visually similar to pizzas while containing the required ingredient. This shows the fine-grain structure of the semantic space where recipes and images are organized by visual or semantic similarity inside the different classes.

    \begin{table*}[t]
        \centering
        \begin{tabular}{c c c c c c}\begin{turn}{90}~~~~~~Carrot\end{turn} &
            \includegraphics[width=0.17\linewidth]{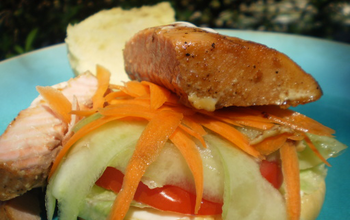} &
            \includegraphics[width=0.17\linewidth]{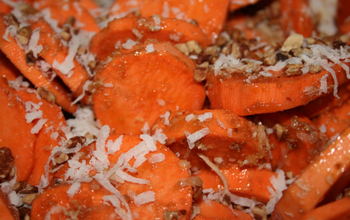} &
            \includegraphics[width=0.17\linewidth]{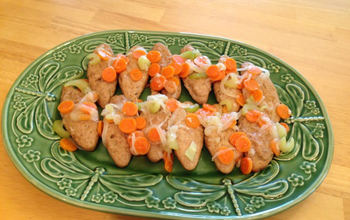} &
            \includegraphics[width=0.17\linewidth]{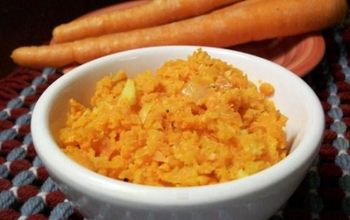} &
            \includegraphics[width=0.17\linewidth]{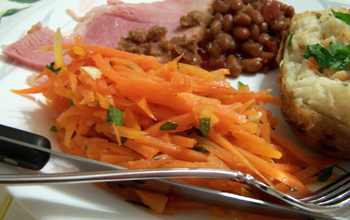} \\
            \begin{turn}{90}~~~Mushroom\end{turn} &
            \includegraphics[width=0.17\linewidth]{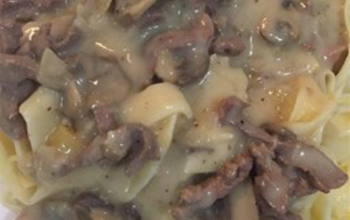} &
            \includegraphics[width=0.17\linewidth]{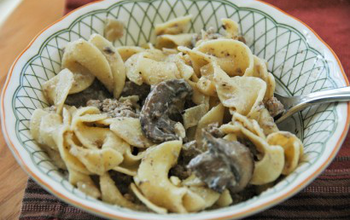} &
            \includegraphics[width=0.17\linewidth]{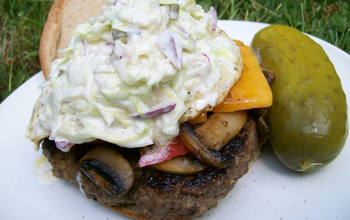} &
            \includegraphics[width=0.17\linewidth]{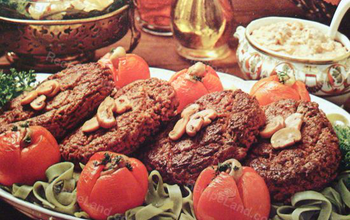} &
            \includegraphics[width=0.17\linewidth]{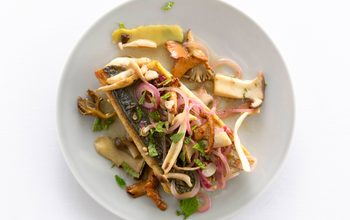}
        \end{tabular}
        \vspace{\baselineskip}
        \vspace{-0.4cm}
        \caption{Examples of images in the top 20 when searching for the ingredient \emph{Carrot} (top row) or \emph{Mushroom} (bottom row).}
        \label{tab:visu_ingr_to_image_mushroom}
        \vspace{-0.6cm}
    \end{table*}

    \begin{table*}[t]
        \centering
        \begin{tabular}{c c c c c}
            
          Mushrooms & Pineapple & Olives & Pepperoni & Strawberries \\
        
          \includegraphics[width=0.17\linewidth]{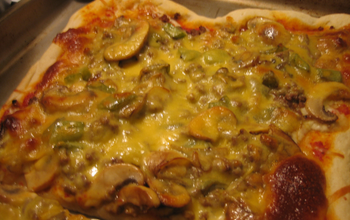} &
          \includegraphics[width=0.17\linewidth]{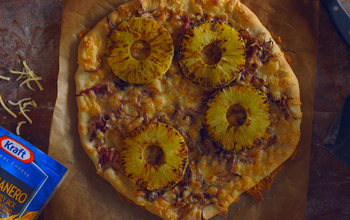} &
          \includegraphics[width=0.17\linewidth]{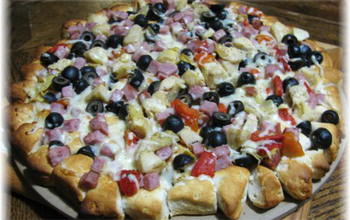} &
          \includegraphics[width=0.17\linewidth]{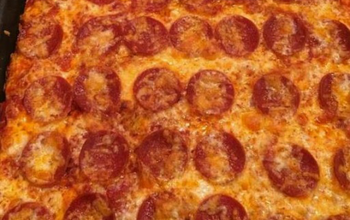} &
          \includegraphics[width=0.17\linewidth]{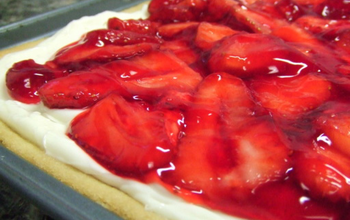} \\
          
          \includegraphics[width=0.17\linewidth]{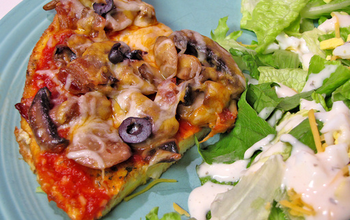} &
          \includegraphics[width=0.17\linewidth]{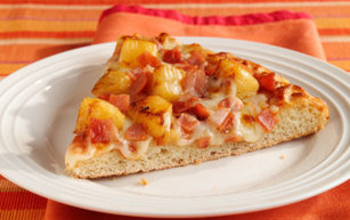} &
          \includegraphics[width=0.17\linewidth]{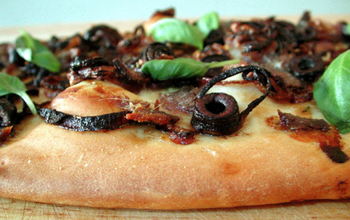} &
          \includegraphics[width=0.17\linewidth]{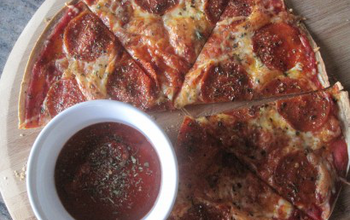} &
          \includegraphics[width=0.17\linewidth]{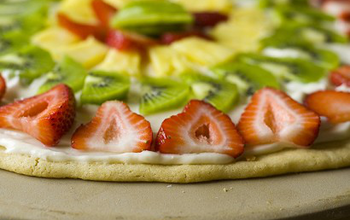}
        \end{tabular}
        \vspace{\baselineskip}
        \vspace{-0.4cm}
        \caption{Examples of images in the top 20 results when searching for a specific ingredient within the class \emph{Pizza}.}
        \label{tab:visu_ingr_to_image_pizza}
        \vspace{-0.6cm}
    \end{table*}

    \subsection{Removing ingredients}

    The capacity of finely model the presence or absence of specific ingredients may be interesting for generating menus, specially for users with dietary restrictions (for instance, peanut or lactose intolerance, or vegetarians and vegans).
    To do so, we randomly select a recipe having \emph{broccoli} in its ingredients list (\autoref{tab:vvisu_remove_ingrs_removed}, first column) and retrieve the top 4 closest images in the embedding space from 1000 recipe images (\autoref{tab:vvisu_remove_ingrs_removed}, top row). Then we remove the \emph{broccoli} in the ingredients and remove the instructions having the \emph{broccoli} word. Finally, we retrieve once again the top 4 images associated to this "modified" recipe (\autoref{tab:vvisu_remove_ingrs_removed}, bottom row).
    
    The retrieved images using the original recipe have broccoli, whereas the retrieved images using the modified recipe do not have broccoli. This reinforces our previous statement, highlighting the ability of our semantic space to correctly discriminate items with respect to ingredients.
    
    \begin{table*}[t]
        \centering
        \begin{tabular}{m{0.17\linewidth}|c c c c c}
            \begin{center}
                Query
            \end{center}
            &
            \multicolumn{4}{c}{Top retrieved images}\\
            
            \hline
            \begin{minipage}{\linewidth}
                \vspace{0.5em}
                \begin{center}\textbf{Tofu Saut\'e}\end{center}
                ~~~~\\
                \small \textit{Oregano, Zucchini, Tofu, Bell pepper, Onions, \textbf{Broccoli}, Olive Oil}\\
                ~~~~\\
                Cut all ingredients into small pieces. Boil water [...]%
            \end{minipage}
            &
            \multicolumn{5}{c}{
                    \begin{minipage}{.68\linewidth}
                        \vspace{0.5em}
                        \includegraphics[width=0.246\linewidth]{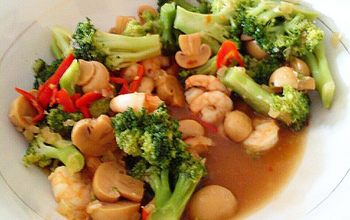}
                        \includegraphics[width=0.246\linewidth]{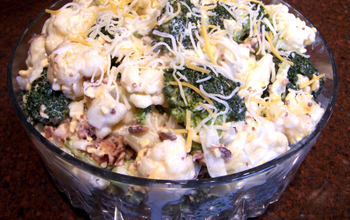}
                        \includegraphics[width=0.246\linewidth]{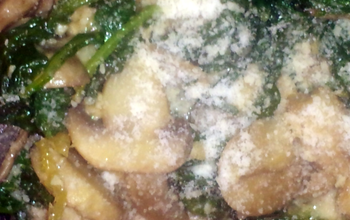}
                        \includegraphics[width=0.246\linewidth]{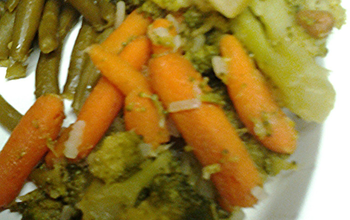} \\
                        \includegraphics[width=0.246\linewidth]{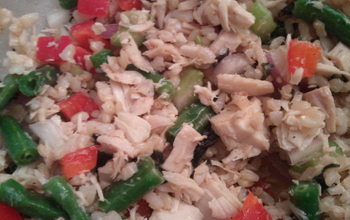}
                        \includegraphics[width=0.246\linewidth]{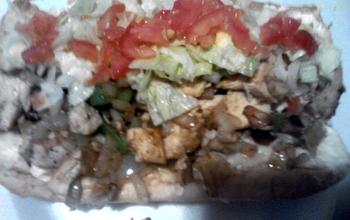}
                        \includegraphics[width=0.246\linewidth]{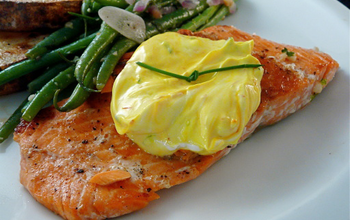}
                        \includegraphics[width=0.246\linewidth]{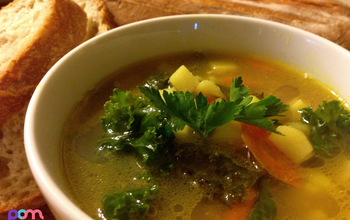}
                    \end{minipage}
            }
        \end{tabular}
        \vspace{\baselineskip}
        \vspace{-0.4cm}
        \caption{Top 4 retrieved images with (top row) and without (bottom row)  broccoli in the ingredient and instruction lists w.r.t the original recipe.}
        \label{tab:vvisu_remove_ingrs_removed}
        \vspace{-0.6cm}
    \end{table*}

    \section{Conclusion}

In this paper, we align cooking recipes and pictures with a metric learning approach. Our method is validated on a large-scale dataset, using the retrieval task (image to text, and vice versa) as the main application.
We show how triplet strategies can attain superior performance when compared to pairwise ones, being able to better align textual and visual information, achieving state-of-the-art results on the Recipe1M dataset.

Then, we discuss the potential of such model on computational cooking applications. Our qualitative studies on downstream tasks (multi-modal retrieval, ingredient to image retrieval, or ingredient removal) demonstrate the efficiency of the semantic space to encode ingredients, instructions, and images. This kind of application opens interesting perspectives for ambitious tasks as menu composition or cooking with restricted ingredient availability.

    \section*{Acknowledgment}
    This work was partially supported by CNPq --- Brazilian's National Council for Scientific and Technological Development --- and by Labex SMART, supported by French state funds managed by the ANR within the Investissements d'Avenir program under reference ANR-11-LABX-65.
    
    \bibliographystyle{IEEEtran}
    
    \bibliography{egbib}

\begin{thebibliography}{10}
\providecommand{\url}[1]{#1}
\csname url@samestyle\endcsname
\providecommand{\newblock}{\relax}
\providecommand{\bibinfo}[2]{#2}
\providecommand{\BIBentrySTDinterwordspacing}{\spaceskip=0pt\relax}
\providecommand{\BIBentryALTinterwordstretchfactor}{4}
\providecommand{\BIBentryALTinterwordspacing}{\spaceskip=\fontdimen2\font plus
\BIBentryALTinterwordstretchfactor\fontdimen3\font minus
  \fontdimen4\font\relax}
\providecommand{\BIBforeignlanguage}[2]{{%
\expandafter\ifx\csname l@#1\endcsname\relax
\typeout{** WARNING: IEEEtran.bst: No hyphenation pattern has been}%
\typeout{** loaded for the language `#1'. Using the pattern for}%
\typeout{** the default language instead.}%
\else
\language=\csname l@#1\endcsname
\fi
#2}}
\providecommand{\BIBdecl}{\relax}
\BIBdecl

\bibitem{Harris54}
Z.~Harris, ``Distributional structure,'' \emph{Word}, vol.~10, no.~23, pp.
  146--162, 1954.

\bibitem{mikolov2013distributed}
T.~Mikolov, I.~Sutskever, K.~Chen, G.~S. Corrado, and J.~Dean, ``Distributed
  representations of words and phrases and their compositionality,'' in
  \emph{NIPS}, 2013, pp. 3111--3119.

\bibitem{chen2016}
J.~Chen and C.-w. Ngo, ``Deep-based ingredient recognition for cooking recipe
  retrieval,'' in \emph{MM}, 2016, pp. 32--41.

\bibitem{wang2015}
X.~Wang, D.~Kumar, N.~Thome, M.~Cord, and F.~Precioso, ``Recipe recognition
  with large multimodal food dataset,'' in \emph{ICMEW}, 2015, pp. 1--6.

\bibitem{SanjoK17}
S.~Sanjo and M.~Katsurai, ``Recipe popularity prediction with deep
  visual-semantic fusion,'' in \emph{CIKM}, 2017, pp. 2279--2282.

\bibitem{salvador2017}
A.~Salvador, N.~Hynes, Y.~Aytar, J.~Marin, F.~Ofli, I.~Weber, and A.~Torralba,
  ``Learning cross-modal embeddings for cooking recipes and food images,'' in
  \emph{CVPR}, 2017.

\bibitem{cea2017}
\emph{CEA2017: Proceedings of the 9th Workshop on Multimedia for Cooking and
  Eating Activities in Conjunction with The 2017 International Joint Conference
  on Artificial Intelligence}, 2017.

\bibitem{chenpfid2009icip}
M.~Chen, K.~Dhingra, W.~Wu, L.~Yang, R.~Sukthankar, and J.~Yang, ``Pfid:
  Pittsburgh fast-food image dataset,'' in \emph{ICIP}, 2009.

\bibitem{Kawano2014}
Y.~Kawano and K.~Yanai, ``Foodcam: A real-time mobile food recognition system
  employing fisher vector,'' in \emph{MMM}, 2014.

\bibitem{Farinella2015}
G.~M. Farinella, D.~Allegra, and F.~Stanco, \emph{A Benchmark Dataset to Study
  the Representation of Food Images}, 2015, pp. 584--599.

\bibitem{bossard2014}
L.~Bossard, M.~Guillaumin, and L.~Van~Gool, ``Food-101 -- mining discriminative
  components with random forests,'' in \emph{ECCV}, 2014.

\bibitem{Beijbom2015}
O.~Beijbom, N.~Joshi, D.~Morris, S.~Saponas, and S.~Khullar, ``Menu-match:
  Restaurant-specific food logging from images,'' in \emph{2015 IEEE Winter
  Conference on Applications of Computer Vision}, 2015.

\bibitem{Goodfellow-et-al-2016}
I.~Goodfellow, Y.~Bengio, and A.~Courville, \emph{Deep Learning}.\hskip 1em
  plus 0.5em minus 0.4em\relax MIT Press, 2016,
  \url{http://www.deeplearningbook.org}.

\bibitem{krizhevsky2012imagenet}
A.~Krizhevsky, I.~Sutskever, and G.~E. Hinton, ``Imagenet classification with
  deep convolutional neural networks,'' in \emph{NIPS}, 2012.

\bibitem{Hochreiter1997}
S.~Hochreiter and J.~Schmidhuber, ``Long short-term memory,'' \emph{Neural
  Comput.}, vol.~9, no.~8, pp. 1735--1780, 1997.

\bibitem{weinbergerlmnn2009}
K.~Q. Weinberger and L.~K. Saul, ``Distance metric learning for large margin
  nearest neighbor classification,'' \emph{J. Mach. Learn. Res.}, vol.~10, pp.
  207--244, 2009.

\bibitem{kiros2015unifying}
R.~Kiros, R.~Salakhutdinov, and R.~S. Zemel, ``Unifying visual-semantic
  embeddings with multimodal neural language models,'' \emph{TACL}, 2015.

\bibitem{karpathy2015deep}
A.~Karpathy and L.~Fei-Fei, ``Deep visual-semantic alignments for generating
  image descriptions,'' in \emph{CVPR}, 2015, pp. 3128--3137.

\bibitem{He2015}
K.~He, X.~Zhang, S.~Ren, and J.~Sun, ``Deep residual learning for image
  recognition,'' \emph{arXiv arXiv:1512.03385}, 2015.

\bibitem{ILSVRC15}
O.~Russakovsky, J.~Deng, H.~Su, J.~Krause, S.~Satheesh, S.~Ma, Z.~Huang,
  A.~Karpathy, A.~Khosla, M.~Bernstein, A.~C. Berg, and L.~Fei-Fei, ``{ImageNet
  Large Scale Visual Recognition Challenge},'' \emph{IJCV}, vol. 115, no.~3,
  pp. 211--252, 2015.

\bibitem{kiros2015}
R.~Kiros, Y.~Zhu, R.~R. Salakhutdinov, R.~Zemel, R.~Urtasun, A.~Torralba, and
  S.~Fidler, ``Skip-thought vectors,'' in \emph{NIPS}, 2015.

\bibitem{hotelling1936relations}
H.~Hotelling, ``Relations between two sets of variates,'' \emph{Biometrika},
  vol.~28, no. 3/4, pp. 321--377, 1936.

\end{thebibliography}

\end{document}